\documentclass{article}

\usepackage[preprint]{neurips_2025}

\usepackage[utf8]{inputenc} 
\usepackage[T1]{fontenc}    
\usepackage{hyperref}       
\usepackage{url}            
\usepackage{booktabs}       
\usepackage{amsfonts}       
\usepackage{nicefrac}       
\usepackage{microtype}      
\usepackage{xcolor}         

\usepackage{graphicx} 
\usepackage{amsmath,amssymb}
\usepackage{bbm}
\usepackage{caption}
\usepackage{subcaption}
\usepackage{float}
\usepackage{multirow}
\usepackage{multicol}
\usepackage{booktabs}
\usepackage{soul}
\usepackage{makecell}

\title{Hypothesis Testing in Imaging Inverse Problems}

\author{%
  $\text{Yiming Xi}^{1,2,3}$ \\
  \And
 $\text{Konstantinos Zygalakis}^{2,3}$ \\
  \And
  $\text{Marcelo Pereyra}^{1,3}$  \\
  \AND
  \vspace{-5mm}
   \\
   1. School of Mathematical and Computer Sciences, Heriot-Watt University \\ 2. School of Mathematics, University of Edinburgh\\ 3. Maxwell Institute for Mathematical Sciences
\AND
  \vspace{-6mm}
   \\
 \{yx2022, m.pereyra\}@hw.ac.uk; k.zygalakis@ed.ac.uk
}

\begin{document}

\maketitle

\begin{abstract}
This paper proposes a framework for semantic hypothesis testing tailored to imaging inverse problems. Modern imaging methods struggle to support hypothesis testing, a core component of the scientific method that is essential for the rigorous interpretation of experiments and robust interfacing with decision-making processes. There are three main reasons why image-based hypothesis testing is challenging. First, the difficulty of using a single observation to simultaneously reconstruct an image, formulate hypotheses, and quantify their statistical significance. Second, the hypotheses encountered in imaging are mostly of semantic nature, rather than quantitative statements about pixel values. Third, it is challenging to control test error probabilities because the null and alternative distributions are often unknown. Our proposed approach addresses these difficulties by leveraging concepts from self-supervised computational imaging, vision-language models, and non-parametric hypothesis testing with e-values. We demonstrate our proposed framework through numerical experiments related to image-based phenotyping, where we achieve excellent power while robustly controlling Type I errors.
\end{abstract}

\section{Introduction}
Digital images inform decisions that have a major impact on the
economy, society, and the environment. Illustrative examples include decisions in medical practice, disaster recovery, agriculture, forestry, climate action, quality control, pollution monitoring, and defence. Such images are often generated by using specialised devices
(e.g., medical scanners, telescopes, and radars), which combine
advanced sensor and instrumentation technology with computational imaging algorithms that transform the acquired data into high-quality images and extract information from them \cite{bhandari2022computational}. More precisely, computational imaging methods seek to perform inference on an unknown image $x_\star$ taking values in a signal set $\mathcal{X} \subset \mathbb{R}^n$, from a measurement $y \in \mathbb{R}^m$ modeled as a realization of \cite{kaipio2006statistical}
\begin{equation}
\label{eq: sampling_distribution}
    Y \sim P(A x_\star)
\end{equation}
where $A \in \mathbb{R}^{n\times m}$ models the deterministic physical aspects of the measurement acquisition process and $P$ is a statistical model describing measurement noise as well as other stochastic aspects of this process. Canonical examples of computational imaging problems include image demosaicing, deblurring, inpainting and tomographic reconstruction (see, e.g., \cite{bhandari2022computational}). Computational imaging problems are usually not well posed, hence there is significant uncertainty about $x_\star$ after observing $y$.

Modern computation imaging methods rely strongly on deep learning techniques to regularize problems and deliver accurate solutions \cite{heckel2025deep, mukherjee2023learned}. Until recently, developing such computational imaging techniques required abundant ground truth data for training, a main drawback for many applications in quantitative and scientific imaging where obtaining reliable ground-truth data can be extremely expensive or even impossible. This has stimulated the development of novel self-supervised strategies that exploit subtle properties in the problem to train models directly from the observed measurement data \cite{chen2023imaging}. Alternatively, many imaging methods rely on pre-trained foundational models \cite{terris2025reconstruct,daras2024survey}.

Despite remarkable progress in image reconstruction accuracy, most computational imaging methods still struggle to reliably quantify the uncertainty in their solutions. Some can deliver accurate confidence regions \cite{pereyra2024equivariant, everink2025self}, but the methodology for supporting more advanced inferences such as hypothesis testing is still in its infancy. This critical methodological gap reduces the value of the reconstructed images as quantitative evidence, hindering the rigorous interpretation of scientific imaging experiments and robust interfacing of imaging pipelines with decision-making processes.

There are three main reasons why endowing computational imaging methods with hypothesis testing capabilities is difficult. First, imaging experiments take place in ephemeral conditions, either because the scene is dynamic (e.g., moving objects), the imaging device is not static (e.g., remote sensing from a drone or satellite), or due to operational constraints (e.g., other demands on the resource). As a result, imaging methods would need to rely on a single measurement for both image reconstruction and hypothesis testing. However, evaluating the statistical significance of a hypothesis with the same data that motivated the hypothesis test is highly non-trivial. Unfortunately, acquiring two independent measurements, one for image reconstruction and one for hypothesis testing, is often unfeasible. 

Second, the hypotheses encountered in imaging problems are mostly of semantic nature, related to visual concepts, rather than quantitative statements about pixel values. Such hypotheses can be specified and communicated in natural language, but they are hard to formulate directly as sets of points in a space of images (e.g., consider modeling the set of color images of size $512\times512$ pixels pertaining to the proposition \emph{``An image of a wheat crop stressed due to dehydration''}).

Third, controlling Type I error probabilities requires knowledge of the distribution of the test statistic under the null hypothesis, which is unknown in practice and difficult to model without resorting to oversimplifying assumptions. In addition, image-based hypothesis tests are usually formulated post-hoc, meaning that methods need to control Type I error probabilities in a zero-shot manner.

This paper presents the following main contributions:
\begin{enumerate}
    \item We propose a novel framework for semantic hypothesis testing in imaging inverse problems. The framework considers hypotheses specified in natural language and tests their statistical significant non-parametrically using the shared image-text representation of a vision-language model (VLM). A noise injection step enables the formulation and testing of the hypotheses on the same measurement data, without compromising statistical guarantees.
    \item We present a theoretical analysis of the proposed framework for the case of a linearised vision-language model encoder and image reconstruction network. This analysis provides clear insights into the main factors underpinning the power of the test.
    
    \item We demonstrate the effectiveness of our framework through a series of numerical experiments where we use a Contrastive Language–Image Pre-training (CLIP) VLM and fully self-supervised image reconstruction networks to implement the approach. We achieve excellent power and Type 1 error control, and consistently outperform testing by zero-shot classification with a temperature-scaled version of the same CLIP model.
\end{enumerate}

\begin{figure}[h!]
    \centering
    \includegraphics[width=0.9\textwidth]{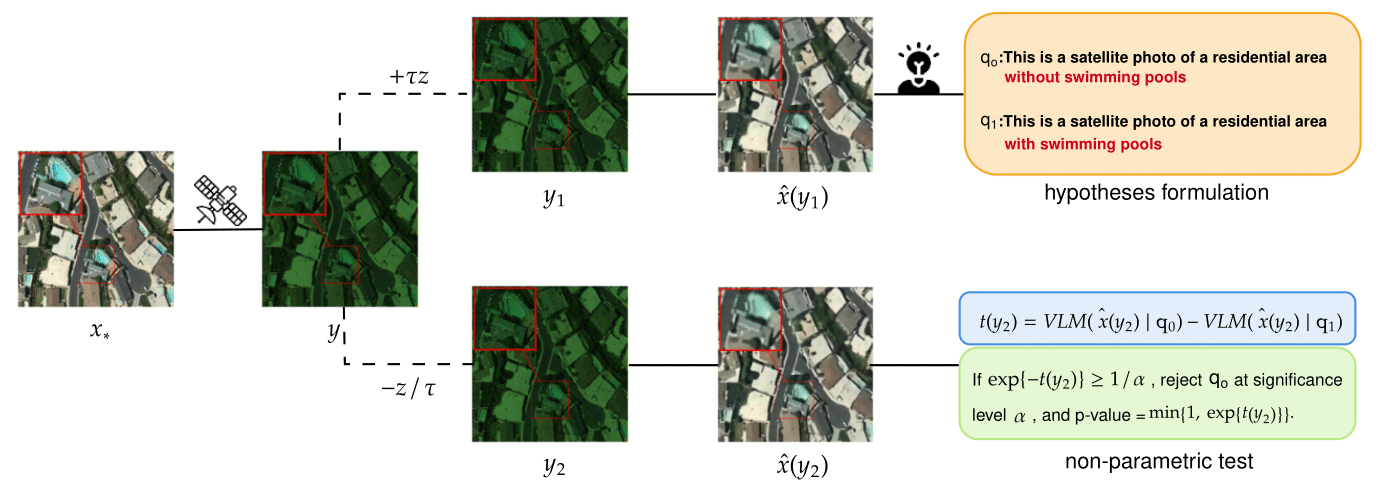}
    \caption{\textbf{Overview of the proposed method.} Given a noisy measurement $y = Ax_\star + \sqrt{\Sigma}\epsilon$, noise-injection generates two new conditionally independent measurements $y_1$ and $y_2$, mimicking experiment replication: $y_1$ is used for image reconstruction and formulation of the hypothesis test, $y_2$ resolves the tests non-parametrically in the shared representation of a VLM and reports a p-value.}
    \label{illustration}
\end{figure}

\section{Background \& Related works}\vspace{-0.3cm} 
\textbf{Hypothesis testing} A statistical hypothesis is a statement about a random variable; e.g., that $Y \sim P(Ax_\star)$ with $x_\star \in \mathcal{X}_0$ for some class of images $\mathcal{X}_0 \subset \mathcal{X}$ (e.g., satellite images of a pine forest). We refer to this statement as the \emph{null} hypothesis $\textrm{H}_0$ when it represents our default position. A hypothesis test is a method of inference used to decide if we should reject $\textrm{H}_0$ in favour of an \emph{alternative} hypothesis $\textrm{H}_1$ based on the observed outcome $Y=y$ \cite{CaseBerg:01}. For example, the alternative $Y \sim P(Ax_\star)$ with $x_\star \in \mathcal{X}_1$ where $\mathcal{X}_1 \subset \mathcal{X}$ and $\mathcal{X}_0 \cap \mathcal{X}_1 = \emptyset$ (e.g., satellite images showing deforestation). Because $Y=y$ contains limited information about $x_\star$, testing procedures can incorrectly reject the null hypothesis, this is known as a Type I error. Similarly, tests may fail to rightfully reject the null hypothesis, a Type II error. Tests are usually calibrated to control the probability of incurring a Type I error under replications of the experiment, the so-called \emph{significance} level. Given two tests with equal significance level, we prefer the test with highest \emph{power} (i.e., the probability of not incurring a Type II error). Moreover, the design of a test requires some assumptions on the distribution of the test statistic. Non-parametric tests make minimal assumptions on this  distribution -they are more robust but potentially less powerful than alternatives with stronger assumptions \cite{Wasserman2007}. 

\textbf{Contrastive Language–Image Pre-training (CLIP)} CLIP is a technique for training multimodal VLMs \cite{radford2021learning}, as well as an essential building block of many other modern foundational models. By using a large amount of image-text pairs, CLIP models learn a shared latent representation for image and text, such that semantically similar image-text pairs are close while dissimilar pairs are far apart. Modern CLIP models, trained on billions of image-text pairs, are remarkably accurate at zero-shot image classification tasks \cite{radford2021learning}. Additionally, they can output approximate probabilities for their classification results via softmax normalization and temperature scaling \cite{Guo2017,minderer2021revisiting}. CLIP models can also be fine-tuned for specific applications by using custom datasets (see, e.g., \cite{Wang2022}). 

\textbf{Bayesian uncertainty quantification in imaging inverse problems.} Uncertainty quantification in imaging problems has been traditionally addressed via Bayesian statistical inference methods \cite{kaipio2006statistical}. A wide range of inference and algorithmic strategies have been studied in the Bayesian imaging literature, from stochastic sampling to optimization and variational approximations \cite{Pereyra2016}. With regards to Bayesian hypothesis testing, we note the optimization-based framework \cite{Pereyra2017,Repetti2019-zv}, which relies on log-concave models and hypotheses that are modelled as convex sets in $\mathcal{X}$. Similar tests are considered in \cite{Durmus2018,Cai2018,Liaudat2024}, again by modelling the hypotheses geometrically in pixel-space. Moreover, \cite{Holden2022} tests for model misspecification in Bayesian imaging models with deep generative priors. Bayesian imaging methods rely increasingly on deep-learning priors to achieve accurate estimation results (see, e.g., \cite{Laumont2022, chung2023diffusion, Coeurdoux2024, daras2024survey, spagnoletti2025latino}). However, designing Bayesian imaging methods that report accurate probabilities is challenging \cite{thong2024}. This has led to the development of other uncertainty quantification frameworks, namely equivariant bootstrapping and self-supervised conformal prediction.

\textbf{Equivariant bootstrapping} Equivariant bootstrapping is a parametric resampling technique designed for
situations in which the set of images $\mathcal{X}$ is invariant to a certain group of transformations (e.g., rotations or small shifts) \cite{pereyra2024equivariant}. By leveraging these symmetries, the equivariant bootstrap reduces the bias stemming from using an estimate $\hat{x}(y)$ of $x_\star$ when approximating the sampling distribution of $Y$. The equivariant bootstrap delivers accurate confidence regions, even when used with a self-supervised estimator trained directly on observed measurement data, bypassing the need for ground truth data.

\textbf{Conformal prediction} Conformal prediction is a powerful statistical framework for constructing uncertainty sets around prediction techniques, such as the estimator $\hat{x}(Y)$ \cite{Angelopoulos2023}. It can be applied to any computational imaging technique to produce sets that are guaranteed to contain the true image with a user-specified probability. Crucially, the sets produced are valid in a distribution-free sense. However, applying conformal prediction to quantitative imaging problems is challenging as it requires abundant ground truth data for calibration, which is often not available. Self-supervised conformal prediction methods avoid this drawback by using Stein risk estimators \cite{everink2025self,amougou2025}.

\textbf{Self-supervised and zero-shot computational imaging techniques.} There are three main deep learning frameworks for developing computational imaging techniques when there is no ground truth data available for training. Equivariant imaging strategies leverage symmetries in the set of images $\mathcal{X}$ and Stein risk estimators to construct end-to-end estimators that can be trained from measurement data alone \cite{chen2023imaging}. Reconstruct-to-reconstruct imaging strategies also learn end-to-end estimators directly from measurement data, by relying on a measurement splitting or noise-injection procedure \cite{monroy2025}. Plug-and-play strategies combine an image denoising algorithm with a physical observation model to iteratively solve computational imaging problems in a zero-shot manner \cite{Kamilov2023,mukherjee2023learned,milanfar2024}.

\section{Methodology}
\subsection{Proposed method}
We are now ready to present our proposed methodology for conducting hypothesis testing in imaging problems. As mentioned previously, our method seeks to address the three key challenges introduced in Section 1, which have traditionally hampered the adoption of hypothesis testing on image data.
\subsubsection{Measurement splitting by noise injection}
First, we adopt a noise-injection measurement splitting step, which bypasses the complications associated with controlling test errors when the test is carried out using the same observation ${y}$ which originally motivated the test. For presentation clarity, we introduce this step for problems involving additive Gaussian noise, and subsequently extend the approach to other noise models. Accordingly, having observed a measurement ${y}$ of ${Y}\sim \mathcal{N}(A{x}_\star, \Sigma)$, we split ${y}$ into two synthetic realization ${y}_1$ and ${y}_2$ that are conditionally independent given ${x}_\star$ by drawing the random variables
\begin{equation}
\label{eq:spit}
    {Y}_1={y}+\tau Z,\quad {Y}_2={y}-\frac{Z}{\tau},
\end{equation}
where $Z \sim \mathcal{N}(0,\Sigma)$ and $\tau > 0$. $Y_1$ and $Y_2$ have mean $A{x}_\star$ and are conditionally independent given ${x}_\star$ for all $\tau >0$, as well as i.i.d. in the case $\tau = 1$ with $Y_1, Y_2 \sim \mathcal{N}(A{x}_\star,2\Sigma)$ (see \cite{monroy2025} for details). 

This simple splitting step emulates the replication of the experiment, at the expense of worsening the signal-to-noise ratio. In doing so, it allows us to utilize the realization $y_1$ to obtain an estimate $\hat{{x}}(y_1)$ of ${x}_\star$ for image reconstruction tasks and the formulation of the hypothesis test, with $y_2$ reserved for safely testing these hypotheses. The parameter $\tau$ controls a trade-off between the accuracy of the estimator $\hat{{x}}(Y_1)$ of ${x}_\star$, and the statistical power of hypothesis tests relying on $Y_2$. This parameter can be fine-tuned for specific experiments when calibration data is available. Otherwise, we find that $\tau = 1$ provides a reliable default value (see the Appendix for a analysis of $\tau$).

\subsubsection{Formulation of semantic hypothesis and test statistic}
As explained previously, the hypotheses encountered when performing statistical testing with image data often pertain to semantic information in the image. Such hypotheses are more accurately described by using written natural language, as opposed to specifying a set of pixel values or images. 

For illustration, consider the following example in remote sensing for natural conservation. Upon observing $\hat{\mathbf{x}}(y_1)$, we seek to determine if $y_2$ favors a proposition $\textrm{q}_0$ related to our default assumption, over an alternative proposition $\textrm{q}_1$ related to deforestation activity, e.g.,
\begin{equation}
    \begin{aligned}
        &\quad \textrm{q}_0: \texttt{This is a satellite photo of a natural grasslands with trees.}\\
        &\quad \textrm{q}_1: \texttt{This is a satellite photo of a farmland with crops.}
    \end{aligned}
\end{equation}
Such propositions $\textrm{q}_0$ and $\textrm{q}_1$, which we will later formalize into the hypotheses in our testing problem, can be easily specified and communicated in natural language. Conversely, it would be hard to directly specify or communicate these hypotheses directly as sets of points in the space of images $\mathcal{X}$.

We propose to tackle image-based hypothesis testing problems by using a test statistic that captures visual concepts in $\hat{x}(y_2)$ and quantifies their degree of semantic agreement with the propositions $\textrm{q}_0$ and $\textrm{q}_1$. More precisely, inspired by the zero-shot image classification capabilities of VLMs \cite{radford2021learning}, we consider a test statistic of the form
\begin{align}\label{eq:statistic_T}
t(Y_2)={D\left(\phi_x(\hat{{x}}(Y_2)),\phi_t(\textrm{q}_0)\right)}-{D\left(\phi_x(\hat{{x}}(Y_2)),\phi_t(\textrm{q}_1)\right)}\,,  
\end{align}
where $\textrm{q}_0$ and $\textrm{q}_1$ are text strings describing in natural language the null and alternative hypotheses, $(\phi_t,\phi_x)$ is a pair of text-image encoders that map text and images to a shared text-image embedding space $\Xi$ in which visual concepts and natural language concepts are aligned, and $D$ is a measure of similarity between two embeddings $u,v \in \Xi$ on this shared representation. Without loss of generality, we herein consider a CLIP model \cite{radford2021learning} where $\Xi$ is a $d$-dimensional hyper-sphere of unit radius and
\begin{equation}\label{eq:temperature}
D(u,v) = \lambda\,{u^\top v}
\end{equation}
where $\lambda >0$ is a so-called temperature hyper-parameter. Accordingly, the statistic $t(Y_2)$ takes positive values when $\hat{{x}}(Y_2)$ is in agreement with the proposition $\textrm{q}_0$, and negative values when $\hat{{x}}(Y_2)$ supports $\textrm{q}_1$. We would recommend using a specialised fine-tune CLIP model if available (see, e.g., \cite{Wang2022}), otherwise one can rely on a foundational CLIP model as default.

\subsubsection{Formulation of the hypotheses and testing via Markov's inequality}
Equipped with \eqref{eq:statistic_T}, to formalize the hypothesis test we must decide the strategy to perform the test. The null distribution for the test statistic \eqref{eq:statistic_T} is often not available in practice, making it challenging to rely on a parametric test. We circumvent this difficulty by considering a non-parametric test based on a so-called e-value, -i.e., a non-negative statistic with expectation bounded by $1$ under the null hypothesis- which allows testing via Markov's inequality \cite{ramdas2025hypothesistestingevalues}. Such tests are potentially much less powerful than parametric alternatives, but they are significantly safer to use.

There are many strategies for expressing \eqref{eq:statistic_T} as an e-value. For ease of interpretation, we use
\begin{align}\label{eq:statistic_E}
E  &=\exp{\{-t(Y_2)\}}\,,
\end{align}
and formalize the null and alternative hypotheses for our testing problem as follows \cite{ramdas2025hypothesistestingevalues} 
\begin{equation}\label{eq:hyp}
\textrm{H}_0: \textbf{E}(E) \leq 1 \,,\quad\text{vs.}\quad \textrm{H}_1: \textbf{E}(E) > 1\,.
\end{equation}
The statistic \eqref{eq:statistic_E} can be seen as a likelihood ratio $p(y_2|\textrm{q}_1)/p(y_2|\textrm{q}_0)$ based on surrogate von Misses-Fisher likelihoods $p(y_2|\textrm{q}_i)\propto\exp{\{\lambda {\phi}_x({\hat{x}}(y_2)) \cdot {\phi}_t(\textrm{q}_i)}\}$ for $i=\{0,1\}$ \cite{Mardia1999}, which are an analogue to Gaussian likelihood ratios for distributions supported on the hyper-sphere. Note that these surrogate likelihoods are only used for the purpose of constructing a test statistic; we do not rely on these models to derive a null distribution to control the level of the test.

Using \eqref{eq:statistic_E}-\eqref{eq:hyp}, we resolve the test by using Markov's inequality. More precisely, because $E$ is non-negative with $\textbf{E}(E) \leq 1$ under $\textrm{H}_0$, we have that 
\begin{align}\label{eq:p-value}
\textbf{P}(E \geq 1/\alpha) \leq \alpha\,,\quad \forall\, \alpha \in (0,1)\,
\end{align}
when $\textrm{H}_0$ holds. Therefore, we reject $\textrm{H}_0$ at significance level $\alpha \in (0,1)$ if $E \geq 1/\alpha$.  Otherwise, we fail to reject $\textrm{H}_0$ at level $\alpha$. In addition, upon observing $Y_2 = y_2$, one can report the test's p-value $p = \textrm{min}(1,1/\varepsilon)$ with $\varepsilon = \exp{\{-\lambda t(y_2)\}}$. However, note that this $p$-variable is conservative, i.e., $\textrm{min}(1,1/E)$ it is not uniform on $[0,1]$ under $\textrm{H}_0$ \cite{ramdas2025hypothesistestingevalues}, so the probability of incorrectly rejecting $\textrm{H}_0$ is often far smaller than $\alpha$. The proposed methodological framework is summarized in Figure 1 with the aide of an illustrative example related to satellite imaging for the detection of illegal structures (the detection of illegal swimming pools carries especial importance in regions with water shortages).

\subsection{Analysis of testing power with a linear image encoder} In order to develop an intuition for the proposed framework, we study the power of our test in the case of a linear image encoder $\Phi \in \mathbb{R}^{d\times n}$ projected on the hyper-sphere, i.e., $\phi(x) = \Phi x / \|\Phi x\|$, and a Gaussian linear observation model
$
Y \sim \mathcal{N}(Ax_\star, {\Sigma})\,,
$
leading to $Y_2 = Ax_\star + \sqrt{\Sigma_\tau}\,\xi$ with $\xi\sim \mathcal{N}(0,\mathbb{I}_m)$ and $\Sigma_\tau = \frac{1+\tau^2}{\tau^2}\Sigma$. From \eqref{eq:p-value}, the probability of rejecting $H_0$ given $x_\star$ is
\begin{equation*}
\begin{split}
    \textrm{Pr}&\left(-t\left(Ax_\star + \sqrt{\Sigma_\tau}\,\xi\right)\geq -\log(\alpha) \right)\quad\\ &\quad\quad\quad\quad\quad\quad\quad=\textrm{Pr}\left(\lambda\Delta_q^\top \frac{\Phi \hat{x}\left(Ax_\star +  \sqrt{\Sigma_\tau}\,\xi\right)}{\|\Phi \hat{x}\left(Ax_\star +  \sqrt{\Sigma_\tau}\,\xi\right)\|}\geq \log(\alpha)\right)\,,
\end{split}
\end{equation*}
where $\Delta_q = \phi_t(\textrm{q}_0) - \phi_t(\textrm{q}_1)$ represents the semantic difference between the null and the alternative hypothesis, as captured by the text encoder $\phi_t$. Moreover, we linearize the estimator $\hat{x}$ at $Ax_\star$, 
\begin{align}
\hat{x}\left(Ax_\star +  \sqrt{\Sigma_\tau}\,\xi\right) &\approx \hat{x}(Ax_\star) + J^\top_{\hat{x}(Ax_\star)} \sqrt{\Sigma_\tau}\,\xi\,,
\\ &\approx x_\star + J^\top_{\hat{x}(Ax_\star)} \sqrt{\Sigma_\tau}\,\xi\,,
\end{align}
where $J_{\hat{x}(Ax_\star)}$ denotes the Jacobian of $\hat{x}$ at $Ax_\star$ and where we assume that the estimator is consistent (i.e., in the noiseless case $\hat{x}(Ax_\star) \approx x_\star$) and that $\|Ax_\star + \sqrt{{\Sigma_\tau}}\xi \| \approx \|Ax_\star\|$. Accordingly, 
\begin{equation}\label{eq:power}
\begin{split}
    \textrm{Pr}&\left(t\left(Ax_\star +  \sqrt{\Sigma_\tau}\,\xi\right)\leq \log(\alpha) \right)\\ &\quad\quad\quad\quad\quad\quad\quad\approx\textrm{Pr}\left(\lambda\Delta_q^\top \frac{\Phi x_\star + \Phi J^\top_{\hat{x}(Ax_\star)} \sqrt{\Sigma_\tau}\,\xi}{\|\Phi x_\star\|}\leq \log(\alpha)\right)\,,\\
 &\quad\quad\quad\quad\quad\quad\quad\approx\textrm{Pr}\left(\Delta_q^\top\frac{\Phi J^\top_{\hat{x}(Ax_\star)}}{{\|\Phi x_\star\|}}\sqrt{{\Sigma_\tau}}\,\xi \leq \frac{\log(\alpha)}{\lambda} - \Delta_q^\top \frac{\Phi x_\star}{\|\Phi x_\star\|}\right)\,.
\end{split}
\end{equation}
From \eqref{eq:power}, we identify two terms that play a key role in the power of the test, in addition to $\alpha$ and $\tau$. On the l.h.s. of the inequality, we observe that the power of the test will be high when $\Delta_q$, which represents the semantic signal that we seek to test, is orthogonal to the leading eigenvectors of the error covariance of $\hat{x}$ on the embedding space $\Xi$ (i.e. the leading eigenvectors of the matrix $(\Phi J^\top_{\hat{x}(Ax_\star)}\sqrt{\Sigma_\tau})^\top (\Phi J^\top_{\hat{x}(Ax_\star)}\sqrt{\Sigma_\tau})$). In lay terms, the test will be more powerful when the signal $\Delta_q$ is not aligned with the uncertainty about $x_\star$. Moreover, the r.h.s. of the inequality reveals that the power increases when $\phi(\textrm{q}_0)$ and $\phi(\textrm{q}_1)$ are equal but with opposite sign (i.e., they represent semantically opposite statements), as this maximises the r.h.s. bound. That is, under $\textrm{H}_1$, we have that $\Phi x_\star / \|\Phi x_\star\| \approx \phi(\textrm{q}_1)$, hence $\Delta_q^\top \Phi x_\star / \|\Phi x_\star\| \approx - 2 \|\phi(\textrm{q}_1)\|$. Furthermore, as expected, we see that the test will have greater power when $\tau$ increases, as this reduces the uncertainty about $x_\star$ (but leaves less information to formulate the test). Of course, the test's power decreases as $\alpha$ vanishes. 

We conclude that, at least in this simplified case, our test will be most effective in situations in which the difference between the propositions $\textrm{q}_0$ and $\textrm{q}_1$ is semantically strong and pertains to visual concepts about $x_\star$ that the estimator $\hat{x}$ can determine from $y_2$ with a low level of uncertainty. More generally, the performance of the test will depend chiefly on the interplay of the embeddings of $\textrm{q}_0$ and $\textrm{q}_1$ on the representation of the VLM, and the ``shape'' of error covariance of $\hat{x}(Y_2)$ on that representation, which depends mainly on the covariance of the measurement noise and the signal support $\mathcal{X}$ (natural images take values close to a low-dimensional manifold in their ambient space).

\subsection{Generalization to other noise distributions from the natural exponential family}
The proposed approach can be straightforwardly extended to other measurement noise distributions from the natural exponential family by using a generalization of the measurement splitting strategy discussed in Section 3.1. We use the generalization proposed recently in \cite{monroy2025} in the context of self-supervised image restoration. The other stages of the proposed methodology remain unchanged. 

In a manner akin \cite{monroy2025}, we now consider measurement models with likelihood functions of the form
\begin{equation}\label{generalized_likelihood}
p(y|x)=h(y)\exp\{y^\top\eta(Ax)-\psi(Ax)\}
\end{equation}

where $A$ is again a known linear measurement operator and $h$, $\eta$ and $\psi$ are known functions satisfying the following properties: $\eta$ is an invertible function, $\psi$ is strictly convex, and $h$ is derived from $\eta$, $\psi$ and $A$ such that $p(y|x)$ is normalized. The class \eqref{generalized_likelihood} includes the Gaussian model considered previously as well as many other noise distributions commonly encountered in imaging problems, such as Poisson, Gamma and Binomial noise. 

Similarly to Section 3.1, having observed $Y=y$, we generate two conditionally independent random variables $Y_1$ and $Y_2$ such that $Y = (1-\beta)Y_1 + \beta Y_2$, both with the same mean $Ax_\star$ as $Y$, and where $\beta \in [0,1]$ is analogous to $\tau$ in \eqref{eq:spit}. We first generate $Y_1$ conditionally to $Y=y$, with density given by
\begin{equation}
p(y_1|y,\beta) = \frac{h_1(y_1)h_2(y-y_1)}{h(y)}
\end{equation}
with 
\begin{align}
h_1(\mathbf{y}_1) &= \int \exp\left(-{s}^\top \mathbf{y}_1 + (1 - \beta) \phi\left( \eta^{-1} \left( \frac{{s}}{1 - \beta} \right) \right) \right) \textrm{d}{s}\,,\\
h_2(\mathbf{y}_2) &= \int \exp\left(-{s}^\top \mathbf{y}_2 + \beta \phi\left( \eta^{-1} \left( \frac{{s}}{\beta} \right) \right) \right) \textrm{d}{s}\,.
\end{align}
Given $Y_1 = y_1$, we simulate $(Y_2|Y=y,Y_1=y_1)$ deterministically by setting $y_2 = \frac{1}{\beta}y-\frac{1-\beta}{\beta}y_1$. 

For example, in the case of the so-called ``scaled'' Poisson noise model often encountered in low-photon imaging, we have $Y = \gamma U$ with $U \sim \textrm{Poi}(Ax/\gamma)$ for some known $\gamma$ that controls the level of shot noise. Having observed $Y=y$, we draw a realization $z$ from $Z \sim \textrm{Bin}(y/\gamma,\beta)$ and set
\begin{equation}
y_1 = \frac{y-\gamma z}{1-\beta}\,,\quad y_2 = \frac{1}{\beta}y-\frac{1-\beta}{\beta}y_1\,.    
\end{equation}

See \cite{monroy2025} for other examples, including Binomial noise as encountered in single-photon imaging and the multiplicative Gamma noise model widely used in synthetic aperture radar and magnetic resonance imaging. For Gaussian noise, we recover the splitting of Section 3.1 by setting $\tau=\sqrt{\beta/1-\beta}$. We present an illustrative example of hypothesis testing in a Poisson imaging problem in the Appendix. 

\subsection{Choice of estimator $\hat{x}$}
In order to implement our framework, we require an estimator $\hat{x}$ to recover $x_\star$ from $Y_1 = y_1$ and $Y_2 = y_2$. To obtain high reconstruction quality and testing performance, as well as mitigate the effect of the noise injection step, we recommend implementing $\hat{x}$ by using a state-of-the-art end-to-end reconstruction network or a state-of-the-art estimator based on a plug-and-play foundational generative model (see, e.g., \cite{spagnoletti2025latino}). One can implement two separate estimators for $Y_1 = y_1$ and  $Y_2 = y_2$, or use a single estimator that is x noise-aware or noise-blind. Our framework is particularly useful when used in combination with estimators that are trained directly from $y_1$ and $y_2$ in a fully self-supervised manner \cite{chen2022robust,monroy2025}. In addition to bypassing the need for ground truth training data, this provides robustness to distribution shift during deployment. Training reconstruction networks in a fully self-supervised manner from scratch requires pooling together many exchangeable imaging problems (in the order of one hundred problems). Alternatively, we recommend implementing our method by using a foundational reconstruction network (see, e.g., \cite{terris2025reconstruct}) and applying self-supervised fine-tuning directly on $y_1$ and $y_2$. Else, we recommend the state-of-the-art reconstruction method \cite{spagnoletti2025latino}, which is training-free and incorporates the imaging operator $A$ and level of noise explicitly.

\section{Experiments}
We demonstrate our proposed framework by applying it to an image-based species phenotyping problem. We consider two datasets, a subset of flower species from Oxford 102 Flower \cite{Nilsback08} and a subset of dog breeds from ImageNet \cite{deng2009imagenet}, known as ImageWoof (see the Appendix for more details). For Oxford 102 Flower, we simulate an aggressive image denoising setting in which $Y \sim N(x_\star,\sigma\mathbb{I})$ with $\sigma=0.5617$ such that the peak-signal-to-noise ratio (PSNR) is 5dB. For ImageWoof, we simulate an extreme image inpainting setting where only a small proportion of image pixels are recorded in order to speed up the acquisition; we set $Y \sim N(A x_\star,\sigma\mathbb{I})$ where $A$ is a random binary mask with independent Bernoulli entries with probability $0.2$ and $\sigma=0.05$, such that the PSNR is 7dB.

\begin{figure}
    \centering
    \begin{subfigure}{0.17\textwidth}
         \centering
         \includegraphics[width=\linewidth]{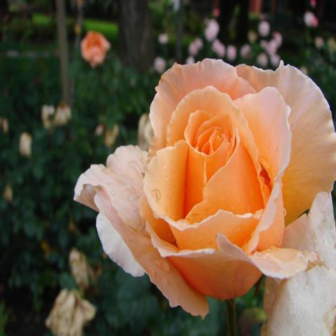}
     \end{subfigure}
              \begin{subfigure}{0.17\textwidth}
         \centering
         \includegraphics[width=\linewidth]{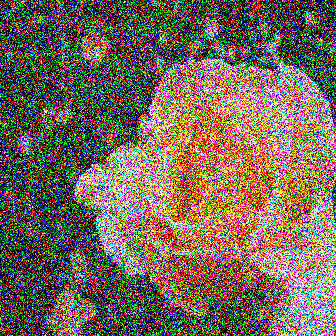}
     \end{subfigure}
         \begin{subfigure}{0.17\textwidth}
         \centering
         \includegraphics[width=\linewidth]{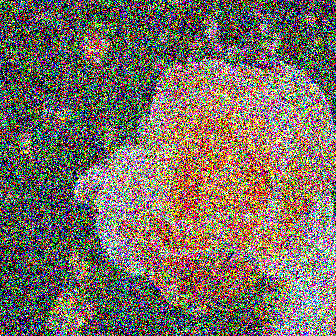}
     \end{subfigure}
         \begin{subfigure}{0.17\textwidth}
         \centering
         \includegraphics[width=\linewidth]{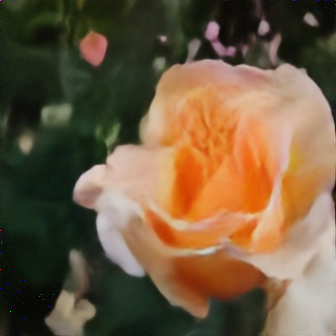}
     \end{subfigure}
         \begin{subfigure}{0.17\textwidth}
         \centering
         \includegraphics[width=\linewidth]{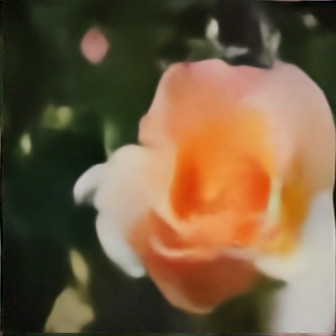}
     \end{subfigure}
    \\

        \begin{subfigure}{0.17\textwidth}
         \centering
         \includegraphics[width=\linewidth]{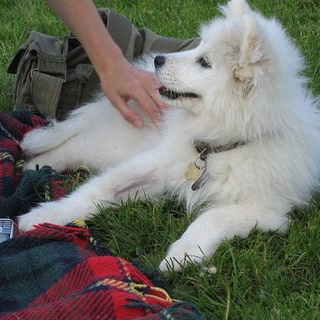}
         \caption{$x_\star$}
     \end{subfigure}
              \begin{subfigure}{0.17\textwidth}
         \centering
         \includegraphics[width=\linewidth]{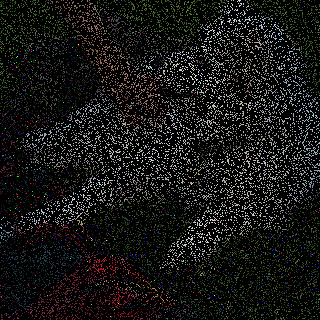}
         \caption{$y$}
     \end{subfigure}
         \begin{subfigure}{0.17\textwidth}
         \centering
         \includegraphics[width=\linewidth]{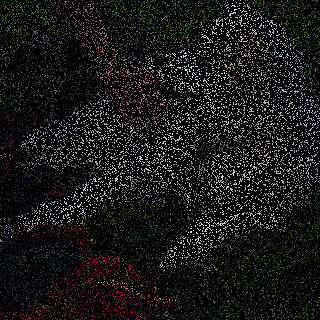}
         \caption{$y_1$}
     \end{subfigure}
         \begin{subfigure}{0.17\textwidth}
         \centering
         \includegraphics[width=\linewidth]{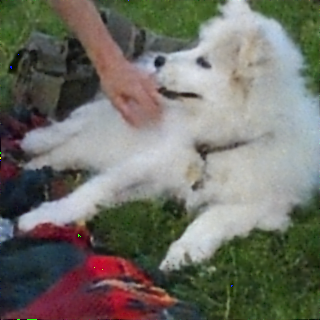}
         \caption{$\hat{x}_{sup}(y_1)$}
     \end{subfigure}
         \begin{subfigure}{0.17\textwidth}
         \centering
         \includegraphics[width=\linewidth]{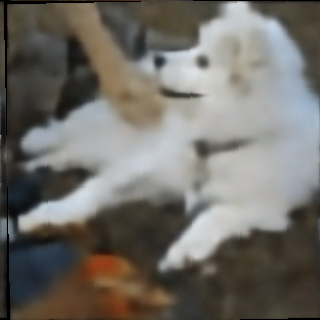}
         \caption{$\hat{x}_{self}(y_1)$}
     \end{subfigure}
     \caption{Example images from Oxford 102 Flower and ImageWoof experiments: true image $x_*$, measurements $y$ and $y_1$, supervised and self-supervised estimates $\hat{x}_{sup}(y_1)$ and $\hat{x}_{self}(y_1)$).}
     \label{ex_recon}
\end{figure}

We implement our method as follows. For the estimator $\hat{x}$, we use a state-of-the-art end-to-end image reconstruction architecture and consider two training strategies. First, for each dataset, we train a fully self-supervised network directly from the measurements $\{y^{(i)}_1\}_{i=1}^M$ obtained by pulling together $M$ exchangeable problems, without any ground truth data. This is achieved by using a Stein estimator of the network's reconstruction mean squared error loss in combination with an equivariant imaging loss \cite{chen2022robust}. For comparison, to assess the impact of self-supervised training on the power of the test, we also train the same network by using a supervised mean squared error loss with truth-measurement pairs $\{x_\star^{(i)}, y^{(i)}_1\}_{i=1}^M$. For illustration, Figure \ref{ex_recon} presents two examples of a true image ${x}_\star$ from the Oxford 102 Flower (top row) and ImageWoof (bottom row) datasets, together with an example of a measurement $y_1$ and the image reconstructions obtained by using the self-supervised and supervised estimators. Moreover, we adjust $\lambda$ by using a small calibration dataset such that $\textbf{E}(E) \lesssim 1$ under $\textrm{H}_0$, where we recall that $E=\exp{\{-t(Y_2)\}}$, and use the default choice $\tau = 1$. Please see the Appendix for more details about the image reconstruction networks used and the calibration of $\lambda$ and $\tau$.

Furthermore, for the definition of the test statistic, we use the CLIP model \cite{radford2021learning}, which is trained on a large-scale collection of publicly available image–text pairs from the internet. To assess the performance of our hypothesis test, we use the labels provided with the Oxford 102 Flower and ImageWoof datasets to measure Type I and Type II error probabilities. For the null hypotheses, we use propositions of the form 
\begin{equation}
 \textrm{q}_0: \texttt{This is a photo of a (flower species / dog breed)}\,.   
\end{equation}
As alternative proposition, if CLIP models could reliably handle negation, we would use \texttt{This is a photo of a (flower / dog) which is \underline{not} a (flower species / dog breed)} as a challenging alternative. However, CLIP models are known to struggle with concept negation \cite{park2025}, raising the possibility that the model would severely misunderstand this proposition as \texttt{This is a photo of a (flower/dog) which \underline{is} a (flower species / dog breed)}. For robustness, we adopt for the simplified alternatively 
\begin{align}
\textrm{q}_1: &\texttt{This is a photo of a (flower / dog) \st{which is not a}}\\
&\texttt{\st{(flower species / dog breed)}}\,,
\end{align}
which is still challenging for the hypothesis testing problem, as it is semantically close to $\textrm{q}_0$.

For comparison, we also report the hypothesis testing results obtained by using the probabilities provided by the same CLIP model when used as a zero-shot classifier. More precisely, we apply a CLIP-based zero-shot classifier to $\hat{x}(y_2)$ with the classes $\textrm{q}_0$ and $\textrm{q}_1$, and obtain probabilities for these classification results via softmax normalisation with temperature scaling \cite{Guo2017}. Accordingly, the class associated with $\textrm{H}_0$ is assigned the probability
\begin{equation}
    p_0(y_2)=\frac{\exp\{D(\phi_x(\hat{{x}}(y_2)),\phi_t(\textrm{q}_0))\}}{\exp\{D(\phi_x(\hat{{x}}(y_2)),\phi_t(\textrm{q}_0))\}+\exp\{D(\phi_x(\hat{{x}}(y_2)),\phi_t(\textrm{q}_1))\}}\,.
\end{equation}
If $p_0(y_2) \leq \alpha$, we reject $\textrm{H}_0$ for $\textrm{H}_1$ at level $\alpha \in (0,1)$.  Otherwise, we fail to reject $\textrm{H}_0$ at this level. Additional details about the CLIP model used and the experiment setup are provided in the Appendix.

\begin{table}
\centering
\caption{Oxford 102 Flowers image denoising experiment: performance results for the temperature-scaled Zero-Shot CLIP and the proposed approaches, with a self-supervised image reconstruction network. For comparison, results with a supervised reconstruction network are also included. The baseline error corresponds to the accuracy of the CLIP model when applied directly to $x_\star$.
}
\begin{tabular}{cccccccc}
\toprule
&&& \multicolumn{5}{c}{Significance Level} \\
\cmidrule{4-8}
&& Baseline & 2\% & 5\% & 10\% & 15\% & 20\% \\
\midrule
\multirow{2}{*}{\makecell{Zero-Shot CLIP \\self-supervised}} & Type I & 1\% & 1.00\% & 1.00\% & 2.00\% & 2.95\% & 3.88\% \\
& Power & - & 46.48\% & 58.22\% & 68.18\% & 73.77\% & 77.48\% \\
\midrule
\multirow{2}{*}{\makecell{Proposed \\self-supervised}} & Type I & 1\% & 1.00\% & 2.00\% & 3.41\% & 4.36\% & 7.06\% \\
& Power & - & 64.34\% & 73.38\% & 79.76\% & 82.98\% & 85.41\% \\
\midrule
\multirow{2}{*}{\makecell{Proposed \\supervised}} & Type I  & 1\% & 1.00\% & 1.00\% & 1.56\% & 1.56\% & 1.56\% \\
& Power & - &  81.35\% &  84.79\% & 87.21\% & 88.62\% & 89.55\% \\
\bottomrule
\end{tabular}
\label{flower}
\end{table}

\begin{table}
\centering
\caption{ImageWoof image inpainting experiment: performance results for the temperature-scaled Zero-Shot CLIP and the proposed approaches, with a self-supervised image reconstruction network. For comparison, results with a supervised reconstruction network are also included. The baseline error corresponds to the accuracy of the CLIP model when applied directly to $x_\star$.}
\begin{tabular}{cccccccc}
\toprule
&&& \multicolumn{5}{c}{Significance Level} \\
\cmidrule{4-8}
&& Baseline & 2\% & 5\% & 10\% & 15\% & 20\% \\
\midrule
\multirow{2}{*}{\makecell{Zero-Shot CLIP \\self-supervised}} & Type I & 4.1\% & 4.26\% & 4.83\% & 5.95\% & 6.84\% & 7.90\% \\
& Power & - & 66.02\% & 76.37\% & 83.79\% & 87.37\% & 89.97\% \\
\midrule
\multirow{2}{*}{\makecell{Proposed \\ self-supervised}} & Type I & 4.1\% & 4.42\% & 4.99\% & 6.11\% & 7.33\% & 3.71\% \\
& Power & - & 74.6\% & 81.54\% & 88.21\% & 90.77\% & 92.93\% \\
\midrule
\multirow{2}{*}{\makecell{Proposed \\supervised}} & Type I & 4.1\% & 4.60\% & 5.08\% & 5.72\% & 6.13\% & 6.79\% \\
& Power & -  & 82.13\% & 88.12\% & 91.86\% & 94.19\% & 95.26\% \\
\bottomrule
\end{tabular}
\label{woof}
\end{table}

Tables \ref{flower} and \ref{woof} summarise the performance of the proposed hypothesis testing approach and the Zero-Shot CLIP method for the two experiments. We report the observed Type I error probabilities and statistical power, at significance levels $\alpha = 0.05, 0.1, 0.15, 0.2$. For comparison, include results for our method when implemented with a supervised image reconstruction network. We also report as baseline the Type I error probability for a CLIP-based zero-shot classifier that observes $x_\star$ directly.

We observe from Tables \ref{flower} and \ref{woof} that the considered CLIP model has a baseline Type I error of the order $1\%$ for the Oxford 102 Flowers dataset and $4\%$ for the ImageWoof dataset. This provides an indication of the accuracy limitations stemming from the epistemic uncertainty of this CLIP model when operated without fine-tuning, regardless of the imaging device and noise conditions. While $4\%$ is a moderately large baseline error for a hypothesis test, we expect that fine-tuning for specific applications and progress in VLM architectures and training will greatly improve this baseline.

Moreover, leaving this baseline error aside, we observe that the zero-shot classifier and the proposed approach are highly conservative, achieving Type I error probabilities that are well below the significance level required (once the baseline is removed). With regards to testing power, relying directly on the probabilities produced by the zero-shot CLIP classifier via softmax normalization generally leads to weak tests. Adopting the proposed approach consistently yields significant improvements in testing power (e.g., an improvement from 58\% to 73\% for the Oxford 102 Flower denoising experiment, at level 5\%). Furthermore, by repeating the experiment with a supervised image reconstruction network, we see that testing power can be significantly further improved by using a more accurate estimator (e.g., a further improvement from 73\% to 84\%, at level 5\%). While relying on a supervised image reconstruction network is not possible in many imaging settings, one can often consider self-supervise fine-tuning of a foundational imaging reconstruction network \cite{terris2025reconstruct} or a zero-shot image reconstruction method derived from a state-of-the-art VLM (e.g., \cite{spagnoletti2025latino}). This approach is illustrated in the Appendix. The Appendix also reports an extensive ablation study.

\section{Discussion and conclusion}
We proposed a novel framework for semantic hypothesis testing in imaging inverse problems, tackling a critical gap in statistical methodology for quantitative computational imaging. The framework uses hypotheses articulated in natural language and assesses their statistical significance non-parametrically and by leveraging the shared image-text representation of a VLM. A key component is a noise injection procedure that enables hypothesis formulation and testing on the same measurement data while preserving statistical validity. We validated the effectiveness of the framework through a series of experiments, employing a CLIP model in conjunction with fully self-supervised image reconstruction networks, thereby eliminating reliance on ground truth data (or a very small amount for calibration). The proposed method demonstrated strong power and rigorous control of Type I error, consistently outperforming zero-shot classification approaches using the same CLIP model. Furthermore, we provided a theoretical analysis in the context of a linearized VLM encoder and image reconstruction network, offering insights into the factors that govern the test's statistical power. 

With regards to limitations of our proposed framework and perspectives for future work. First, the noise-injection step decreases the accuracy of the reconstructions, so future work should develop strategies to automatically optimize $\tau$. Second, while a fully parametric testing approach is likely unfeasible, a better understanding of the null distributions involved would enable more powerful tests through better e-variables and e-to-p calibrators \cite{ramdas2025hypothesistestingevalues}. Third, CLIP models and other existing VLMs struggle with negation and suffer from biases, diminishing the reliability of our framework. We hope and anticipate that these issues will be mitigated by progress and democratization of VLM technology. Moreover, our framework currently operates a single test, it should be extended to multiple testing problems. Furthermore, future work should also also develop better strategies to set $\lambda$ automatically (e.g., by maximum likelihood estimation). Lastly, our framework assumes that the imaging forward model is perfectly known, it should be extended to handle partially unknown models and equipped with safeguards to account for modelling errors. We believe that, with these extensions, the proposed framework would be potentially widely adopted across many impactful application domains that rely on quantitative and scientific images to inform high-value decisions and conclusions.

\section{Acknowledgments and Disclosure of Funding}
We are grateful to Andres Almansa, Alain Durmus, Pierre Chainais, Tobias Liaudat and Julian Tachella for useful discussion. This work was supported by the UK's Engineering and Physical Sciences Research Council (EPSRC) under grants EP/Z534481/1, EP/V006134/1 and EP/S023291/1. We acknowledge the use of the HWU high-performance computing facility (DMOG) and associated  support services in the completion of this work.

\newpage
\bibliographystyle{plainnat}
\bibliography{ref.bib}

\newpage
\appendix
\section{Experimental Details}
\subsection{Model and training details}
\paragraph{Architecture}
For our image reconstruction network $\hat{x}(\cdot)$ we follow \cite{pereyra2024equivariant} and use an architecture based on an unrolled proximal half-quadratic splitting algorithm with three iterations, where the proximal operator is replaced by a trainable U-Net architecture featuring four scales. A single shared proximal network is employed across all iterations (i.e., with tied weights). The resulting unrolled network comprises 8554375 trainable parameters. The supervised network is trained by using a mean squared error (MSE) loss, while for the self-supervised network, we apply the robust equivariant imaging loss \cite{chen2021equivariant} with random horizontal and vertical pixel shifts as transformations, which also optimizes MSE performance but achieves this directly from the measurement data.

\paragraph{Optimization and Schedule}
We trained our neural networks using Adam optimizer with learning rate $\eta=3\times10^{-5}$. To further refine the training process, a step-based learning rate scheduler was employed, which reduces the learning rate by a factor of 0.3 every 10 training epochs.

\paragraph{Resources used}
All experiments were conducted with the open-source \texttt{DeepInverse} library \cite{tachella2023deepinverse} on a high-performance computing cluster equipped with NVIDIA A40 GPUs (48 GB), AMD EPYC 7543 processors with 32 cores, and 256 GB of system memory. Training each reconstruction network required approximately 3 GPU days. Inference required in the order of 3.5 seconds per image.

\subsection{Dataset}
The Oxford 102 Flowers dataset comprises 102 flower categories, with each category containing between 40 and 258 images. We split each the dataset into 80\% for training and 20\% for testing, across all categories. ImageWoof, a subset of 10 visually similar dog breeds from ImageNet, includes approximately 900 training images and 400 testing images per class. We train our models on the training sets and perform inference on the corresponding test sets. We report the average reconstruction PSNR for our reconstruction networks here in Table \ref{recon}.

\begin{table}[H]
    \centering
        \caption{Reconstruction performance (PSNR, in dB units) for the supervised and self-supervised networks on the Oxford 102 Flowers dataset and the ImageWoof dataset.}
    \begin{tabular}[t]{ccc}
        \toprule
       & Oxford Flowers&ImageWoof\\
        \cmidrule{2-3}
        supervised& 24.6 & 28.3\\
        self-supervised& 24.2 & 25.5\\
    \bottomrule
    \end{tabular}
    \label{recon}
\end{table}

\section{CLIP}
For our hypothesis testing inference tasks, we used the original CLIP model developed by OpenAI \cite{radford2021learning} without any additional fine-tuning. CLIP (Contrastive Language–Image Pre-training) is a neural network that learns visual concepts from natural language supervision. After leveraging a Vision Transformer and a Text Transformer, CLIP efficiently achieves great zero shot performance by adopting a contrastive objective aligning image and text embeddings. Given an image and a list of text prompts, CLIP computes a cosine similarity score between the image and each prompt. These scores can either be used directly, e.g., for zero-shot classification, or transformed into probabilities by temperature-scaled softmax normalization and propagated to downstream tasks. Following the original model, several domain-specific variants have been introduced to better handle specialized data. For example, RemoteCLIP for satellite images \cite{liu2024remoteclip} and MedCLIP for medical imaging tasks \cite{Wang2022}.

\section{Hyperparameter tuning}
For each experiment and network, we calibrate the temperature hyperparameter $\lambda$ with a small calibration dataset such that $\textbf{E}(E)$ is as close to $0.98$ under null distribution. We summarize the choice of $\lambda$ in Table \ref{lambda}. 

\begin{table}[h]
    \centering
        \caption{Temperature hyperparameter $\lambda$ for the supervised and self-supervised reconstruction networks across datasets, adjusted such that $\textbf{E}(E) \approx 0.98$ under the null distribution.}
    \begin{tabular}[t]{ccc}
        \toprule
       & Oxford Flowers&ImageWoof\\
        \cmidrule{2-3}
        supervised& 3.05 & 1.23\\
        unsupervised& 1.44 & 1.12\\
    \bottomrule
    \end{tabular}
    \label{lambda}
\end{table}
\section{Illustrative example - Poisson image denoising experiments}

In order to illustrate the effectiveness of the proposed framework in situations involving non-Gaussian noise, we now consider a low-photon Poisson image denoising problem with shot-noise parameter $\gamma=0.5$. We use the Oxford 102 Flowers dataset, apply the appropriate noise-injection step (16), and set $\beta=0.15$ as recommended in \cite{monroy2025}. With regards to the reconstruction network, we use the same architecture as previously in combination with a self-supervised training approach for Poisson data based on a generalized recorrupted-to-recorrupted loss \cite{monroy2025}. Again, this approach is designed to optimize MSE performance directly from measurement data, bypassing the need for clean reference images for training. The average reconstruction PSNR for this experiment is 23.9dB, which is high given the difficulty of the problem. Table \ref{poi_result} summarizes the Type I error probability and statistical power results for this experiment (we used $\lambda=1.44$). For completeness, Figure \ref{poi_flower} below depicts a set of example images for this problem. Again, we observe that our proposed framework provides excellent control of Type I error probabilities and power.

\begin{figure}[h]
\centering

        \begin{subfigure}{0.23\textwidth}
         \centering
         \includegraphics[width=\linewidth]{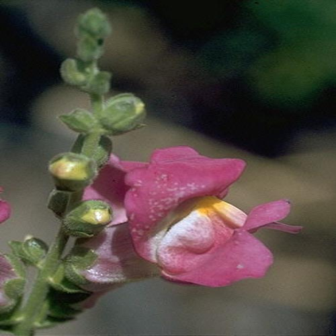}
         \caption{$x_\star$}
     \end{subfigure}
              \begin{subfigure}{0.23\textwidth}
         \centering
         \includegraphics[width=\linewidth]{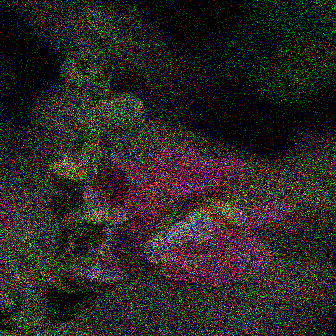}
         \caption{$y$}
     \end{subfigure}
         \begin{subfigure}{0.23\textwidth}
         \centering
         \includegraphics[width=\linewidth]{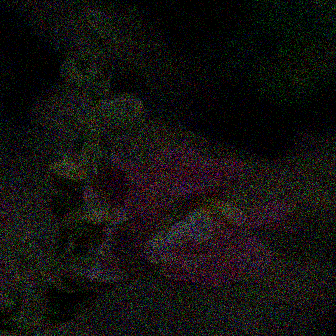}
         \caption{$y_1$}
     \end{subfigure}
         \begin{subfigure}{0.23\textwidth}
         \centering
         \includegraphics[width=\linewidth]{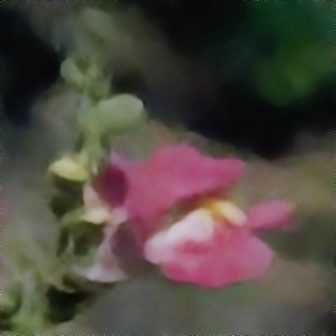}
         \caption{$\hat{x}(y_1)$}
     \end{subfigure}
     \caption{Example images from Oxford 102 Flower with Poisson noise: true image $x_*$, measurements $y$ and $y_1$, self-supervised estimates $\hat{x}(y_1)$.}
     \label{poi_flower}
\end{figure}

\begin{table}
\centering
\caption{Oxford 102 Flowers image Poisson denoising experiment: performance results with a self-supervised image reconstruction network.}
\begin{tabular}{cccccccc}
\toprule
&&& \multicolumn{5}{c}{Significance Level} \\
\cmidrule{4-8}
&&Baseline&  2\% & 5\% & 10\% & 15\% & 20\% \\
\midrule

\multirow{2}{*}{\makecell{Proposed\\(PSNR 23.90dB)}} & Type I &1\%& 1.56\%  & 1.56\% & 1.83\% & 3.39\% & 3.39\% \\
& Power &-& 81.85\% & 86.43\% & 89.38\% & 90.77\% & 92.04\% \\

\bottomrule
\end{tabular}
\label{poi_result}
\end{table}
\section{Ablation Study}
\subsection{Analysis of the parameter $\tau$}
\begin{figure}
    \centering
    \includegraphics[width=0.8\textwidth]{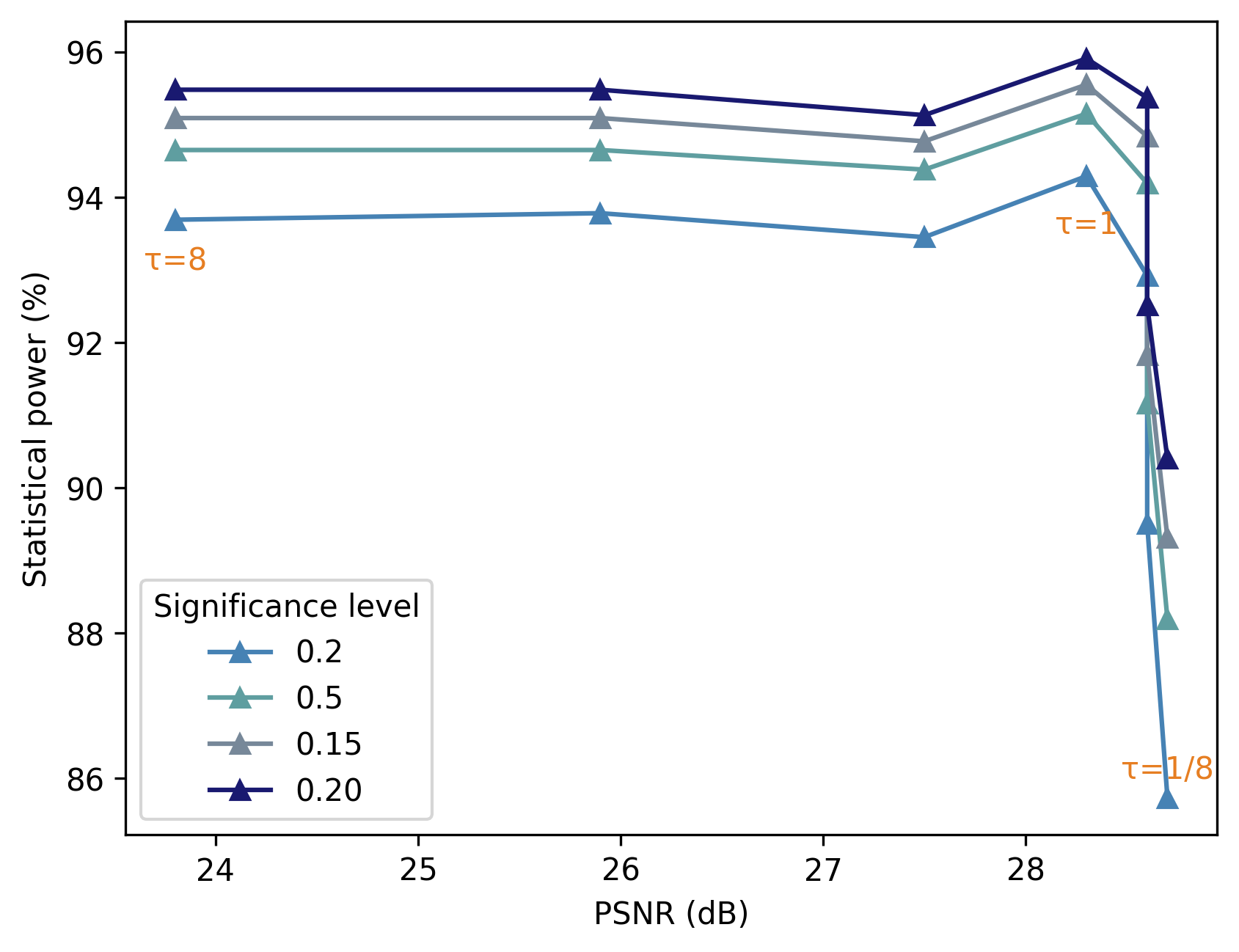}
    \caption{Trade-off between reconstruction quality and statistical power for different values of $\tau$. The x-axis shows the PSNR of $y_1$, while the y-axis represents the statistical power obtained from $y_2$ at different significance levels. PSNR values are labeled according to the corresponding values of $\tau$.}
    \label{tau}
\end{figure}

The noise-injection step (2) allows splitting the original measurement into two independent measurements with lower signal-to-noise ratio. In doing so, it enables hypothesis testing but reduces reconstruction accuracy. The parameter $\tau$ controls the trade-off between the accuracy of the reconstruction obtained from $y_1$ and the statistical power of the test based on $y_2$. 

Figure \ref{tau} illustrates this trade-off in the case of ImageWoof image inpainting experiment with supervised reconstruction network, for a range of values $\tau$ from $\tau = 1/8$ to $\tau = 8$. We observe that taking $\tau \geq 1$ provides excellent power, with $\tau = 1$ providing the best balance of reconstruction quality and statistical power. Taking $\tau \gg 1$ significantly reduces reconstruction quality without gains in power, whereas $\tau < 1$ marginally improves reconstruction quality w.r.t. $\tau = 1$ but leads to a significant deterioration of power. We recommend using $\tau = 1$.

\subsection{Variability of statistical power results}
The statistical power results reported in Tables 1 and 2 correspond to mean values. Repetitions of the experiments produce similar results with some small variability in the observed statistical power. This variability has a standard deviation of the order of $0.2\%$ for Table 1 and $0.3\%$ for Table 2.

\subsection{Using a reconstruction network derived from a foundational image reconstruction model}
\begin{table}
\centering
\caption{ImageWoof image inpainting experiment: Type I error probability and statistical power for the proposed framework implemented with the RAM \cite{terris2025reconstruct} reconstruction network (zero-shot and with self-supervised fine-tuning). }
\begin{tabular}{cccccccc}
\toprule
&&& \multicolumn{5}{c}{Significance Level} \\
\cmidrule{4-8}
&&Baseline&  2\% & 5\% & 10\% & 15\% & 20\% \\
\midrule

\multirow{2}{*}{\makecell{zero-shot RAM\\ (PSNR 24.67dB)}} & Type I &4.1\%& 4.58\%  & 5.74\% & 6.39\% & 6.96\% & 7.53\% \\
& Power &- &76.61\% & 83.65\% & 88.32\% & 91.04\% & 92.46\% \\
\midrule
\multirow{2}{*}{\makecell{w. finetuning \\ (PSNR 27.20dB)}} & Type I &4.1\%& 4.59\%  & 4.92\% & 5.41\% & 6.56\% & 6.97
\% \\ 
& Power &-&  82.97\%& 88.67\% & 91.45\% & 92.98\% & 94.37\%  \\

\bottomrule
\end{tabular}
\label{ram}
\end{table}

As an alternative to the supervised and self-supervised reconstruction networks considered previously, we now implement our framework by using the so-called reconstruct anything model (RAM) foundational reconstruction network \cite{terris2025reconstruct}, which was developed concurrently with our framework. We consider two cases, RAM (zero-shot) and RAM with self-supervised fine-tuning via a robust equivariant imaging loss (fine-tuning RAM requires 1.5 GPU hours in our case with 4000 images). Table \ref{ram} summarizes the Type I and statistical power results obtained with these reconstruction networks for the ImageWoof image inpainting experiment (we use $\lambda = 1.7$). We observe that RAM with fine-tuning achieves excellent results, comparable to the results obtained with the supervised network (see Table 2), without the need for ground truth data, and with a significantly lower computational cost than self-supervised training from scratch.

\subsection{An alternative to testing with e-values: Equivariant bootstrapping and sign test}

Our proposed framework formulates the hypothesis test by using an e-value approach combined with Markov's inequality. An interesting alternative, which also allows performing the test non-parametrically, is to use a bootstrapping algorithm in combination with a sign test. To explore this alternative, we leverage the recently proposed equivariant bootstrapping algorithm \cite{pereyra2024equivariant} which exploits symmetries in the signal set $\mathcal{X}$ in order to provide an accurate Monte Carlo approximation of the sampling distribution of $\hat{x}(Y)$ -we denote these Monte Carlo samples by $\{\tilde{x}_k\}_{k=1}^K$. To the best of our knowledge, this is the most accurate resampling technique currently available for the class of computational imaging problems considered here. To formulate the test, for each element of the sample $\{\tilde{x}_k\}_{k=1}^K$, we compute the test statistic
\begin{equation*}
\tilde{t}_k(Y_2)=\log\frac{{D\left(\phi_x(\tilde{x}_k),\phi_t(\textrm{q}_0)\right)}}{{D\left(\phi_x(\tilde{x}_k),\phi_t(\textrm{q}_1)\right)}}\,,    
\end{equation*}
which takes positive values when $\tilde{x}_k$ is semantically close to $\textrm{q}_0$ and negative values when $\tilde{x}_k$ is semantically close to $\textrm{q}_1$, and where we formalize $\textrm{H}_0$ and $\textrm{H}_1$ as follows
\begin{equation}
    \begin{aligned}
    &\widetilde{H}_0: \text{median}(\tilde{t}_k)\geq \kappa\\
    &\widetilde{H}_1: \text{median}(\tilde{t}_k) < \kappa\,,
\end{aligned}
\end{equation}
for some $\kappa > 0$ that can be adjusted by using a calibration dataset. Equipped with the bootstrap Monte Carlo sample $\{\tilde{t}_k(Y_2)\}$, we straightforwardly resolve the test by using a sign test \cite{dixon1946statistical}, which makes minimal assumptions on the sampling distribution. 

Table \ref{test_eqb} below summarizes the performance of this variant of the proposed framework; the corresponding values of $\kappa$ are reported in Table \ref{kappa}. These results are obtained by using an equivariant bootstrap algorithm that relies on random vertical and horizontal pixel shifts in order to reduce bootstrapping bias. For the ImageWoof dataset, we observe that the statistical power of the framework with the sign test is consistently lower than that of the proposed framework when using e-values. In terms of Type I error, the two variants of the framework are comparable, the variant with the bootstrap and sign test performs better with a supervised network but worse with an unsupervised network. The threshold hyperparameter $\kappa$ can improve Type I error control, but at the expense of reduced statistical power. Moreover, for the Oxford 102 Flowers dataset, we observe that the variant with bootstrap and sign test yields competitive statistical power for both the supervised and unsupervised networks, but with worse control of Type I errors, as removing the bias from bootstrapping is more challenging in this case (this is related to the fact that the error is spread across pixels, as opposed to being concentrated close to a subspace, see \cite{pereyra2024equivariant} for details).

\begin{table}
\centering
\caption{$\kappa$ for different set of experiments being 1\% of the true median value.} 
\begin{tabular}{ccc}
\toprule
            & Oxford Flowers & ImageWoof \\
\cmidrule{2-3}
supervised & -0.030 & -0.170\\
\midrule
unsupervised  & -0.060  &-0.165\\
\bottomrule
\end{tabular}
\label{kappa}
\end{table}

\begin{table}
\centering
\caption{Statistical test results of supervised and unsupervised networks on Oxford Flowers and ImageWoof datasets using equivariant boostrap sampling method with tuned $\kappa$.}
\resizebox{\textwidth}{!}{%
\begin{tabular}{ccccccccc}
\toprule
&&&& \multicolumn{5}{c}{Significance Level} \\
\cmidrule{5-9}
&&&Baseline& 2\% & 5\% & 10\% & 15\% & 20\% \\
\midrule
\multirow{4}{*}{Oxford Flowers} & \multirow{2}{*}{Supervised} & Type I & 1\%& 1.99\% & 2.69\% & 2.69\% & 3.30\% & 3.87\% \\
&& Power &-& 87.24\% & 89.36\% & 89.36\% & 90.37\% & 91.31\% \\
& \multirow{2}{*}{Unsupervised} & Type I &1\%& 2.46\% & 2.88\% & 2.88\% & 3.01\% & 3.31\% \\
&& Power &-& 82.78\% & 85.19\% & 85.19\% & 86.30\% & 87.48\% \\
\midrule

\multirow{4}{*}{ImageWoof} & \multirow{2}{*}{Supervised} & Type I &4.1\%& 4.10\% & 4.14\% & 4.14\% & 4.15\% & 4.16\% \\
&& Power &-& 81.71\% & 83.99\% & 83.99\% & 85.01\% & 85.91\% \\
& \multirow{2}{*}{Unsupervised} & Type I &4.1\%& 4.13\% & 4.16\% & 4.16\% & 4.18\% & 4.18\% \\
&& Power &-& 59.91\% & 63.13\% & 63.13\% & 64.8\% & 66.43\% \\
\bottomrule
\end{tabular}}
\label{test_eqb}
\end{table}

\section{Limitations and Broader Impacts}
\paragraph{Limitations} As mentioned in Section 5, our proposed framework has some limitations that future work should seek to address. The noise-injection step decreases the accuracy of the reconstructions, limiting the amount of information that can extracted from the data. Future work should develop strategies to minimize this drawback by 1) optimizing the parameter $\tau$ which controls the trade-off between reconstruction quality and hypothesis testing power, and 2) through better e-variables and e-to-p calibrators \cite{ramdas2025hypothesistestingevalues} that leads to more powerful tests. Moreover, for simplicity, our framework currently operates a single test. Future work should extend the framework by leveraging the vast recent literature on multiple hypothesis testing. Furthermore, our framework currently assumes that the imaging forward model is fully known, an assumption that is difficult to verify in practice. Future work should increase the robustness of our framework by 1) incorporating the capacity to deal with semi-blind imaging problems involving partially unknown forward models; 2) through better strategies to estimate $\lambda$, and 3) via techniques from robust statistics that for testing with imperfect models \cite{huber1965robust, levy2008robust, gul2017minimax, gao2018robust}.

\paragraph{Broader impacts}
Computational imaging research generally poses some ethical dilemmas, especially when it involves significant methodological innovation. Although the applications studied in the project do not raise any direct concerns, we are aware that the proposed quantitative imaging framework could be adapted and transferred towards military applications by a hostile actor and for violating citizen privacy, for example. In addition, our framework relies on VLM technology that currently suffers from biases, which could lead to leading to unfair or discriminatory outcomes. We hope that these critical issues will be resolved by progress and democratization of VLM technology.

Moreover, the machine learning facet of the proposed research falls within the “limited risk” level of the EU AI Act. We have followed the Act’s recommendations for this level, namely a focus on the full transparency of the training data and the methods used. This also links in with our open research philosophy - we will publicly available release open-source implementations for research codes, with clear documentation and notebook demonstrations. We see this as an essential requirement for conducting research in a democratic knowledge society.

\end{document}